\documentclass[manuscript,acmsmall]{acmart}

\AtBeginDocument{%
  \providecommand\BibTeX{{%
    \normalfont B\kern-0.5em{\scshape i\kern-0.25em b}\kern-0.8em\TeX}}}

\usepackage{times}
\usepackage{epsfig}
\usepackage{graphicx}
\usepackage{amsmath}
\usepackage{multirow}
\usepackage{diagbox}
\usepackage[normalem]{ulem} 
\usepackage{tabularx}
\usepackage{xcolor}
\usepackage{url}
\usepackage{xspace}
\usepackage{caption}
\usepackage{subcaption}
\usepackage{array}
\usepackage{makecell}
\usepackage{stfloats}

\definecolor{RH}{HTML}{ff0000}
\definecolor{LZ}{HTML}{ff9900}
\definecolor{YG}{HTML}{6aa84f}
\definecolor{JW}{HTML}{2476b3}
\definecolor{RB}{HTML}{5476b3}
\definecolor{spare}{HTML}{1c4587}

\newcommand{\technique}{GRADIA\xspace}
\newcommand{\framework}{IAA\xspace}
\newcommand{\frameworkfull}{Interactive Attention Alignment\xspace}

\begin{document}
\title[Eyes between Humans and Deep Neural Network]{Aligning Eyes between Humans and Deep Neural Network through Interactive Attention Alignment}

  
\author{Yuyang Gao}
\email{yuyang.gao@emory.edu}
\affiliation{%
 \institution{Emory University}
 \country{USA}
}

\author{Tong Sun}
\email{tsun8@gmu.edu}
\affiliation{%
 \institution{George Mason University}
 \country{USA}
}

\author{Liang Zhao}
\email{liang.zhao@emory.edu}
\affiliation{%
 \institution{Emory University}
 \state{Georgia}
 \country{USA}
}

\author{Sungsoo Hong}
\email{shong31@gmu.edu}
\affiliation{%
 \institution{George Mason University}
 \country{USA}
}

%
\renewcommand{\shortauthors}{Gao, et al.}

\begin{abstract}
While Deep Neural Networks (DNNs) are deriving the major innovations in nearly every field through their powerful automation, we are also witnessing the peril behind automation as a form of \textit{bias}, such as automated racism, gender bias, and adversarial bias. 
As the societal impact of DNNs grows, finding an effective way to steer DNNs to align their behavior with the human mental model has become indispensable in realizing fair and accountable models.
We propose a novel framework of Interactive Attention Alignment (IAA) that aims at realizing human-steerable Deep Neural Networks (DNNs). IAA leverages DNN model explanation method as an interactive medium that humans can use to unveil the cases of biased model attention and directly adjust the attention. In improving the DNN using human-generated adjusted attention, we introduce GRADIA, a novel computational pipeline that jointly maximizes attention quality and prediction accuracy. We evaluated IAA framework in Study 1 and GRADIA in Study 2 in a gender classification problem. Study 1 found applying IAA can significantly improve the perceived quality of model attention from human eyes. In Study 2, we found using GRADIA can (1) significantly improve the perceived quality of model attention and (2) significantly improve model performance in scenarios where the training samples are limited. We present implications for future interactive user interfaces design towards human-alignable AI.
\end{abstract}


\maketitle

\section{Introduction}

Deep Neural networks (DNNs) are becoming the powerhouse of innovation in our society; they drive vehicles on behalf of humans~\cite{Tesla2020Auto}, replace repetitive tasks in banking and finance industry~\cite{hong2020human}, and provide powerful support for border control agencies to boost security~\cite{John2018How}.
While we are witnessing how DNNs' powerful automation can benefit humanity, at the same time, we are experiencing the peril behind the automation as a form of \textit{bias} in several areas~\cite{Feng2019The}.
As understanding how DNNs behave is the first step to remove the bias and retain the model fairness and accountability, recent years have seen an explosion of interests in model interpretability~\cite{gil201920, holstein2019improving} and understanding when models can fail~\cite{guidotti2019survey}.
However, DNNs offer only limited transparency about underling logical structure for prediction compared to other white-box models, imposing challenges when humans attempt to (1) determine DNN's logical structure to understand when the biased outcomes could occur~\cite{hong2020human} and (2) debug the model to make it align with human expectations~\cite{holstein2019improving, hohman2018visual}.

\begin{table*}[]
\centering
\begin{tabular}{l|ll}
\hline
                               & \textbf{Reasonable Attention} & \textbf{Unreasonable Attention} \\ \hline
\textbf{Accuracte Prediction}  & \textbf{RA:} Reasonable Accurate & \textbf{UA:} Unreasonable Accurate \\
\textbf{Inaccurate Prediction} & \textbf{RIA:} Reasonable Inaccurate & \textbf{UIA:} Unreasonable Inaccurate \\ \hline
\end{tabular}
\end{table*}
\begin{figure*}[]
   \centering
   \includegraphics[width=1\textwidth]{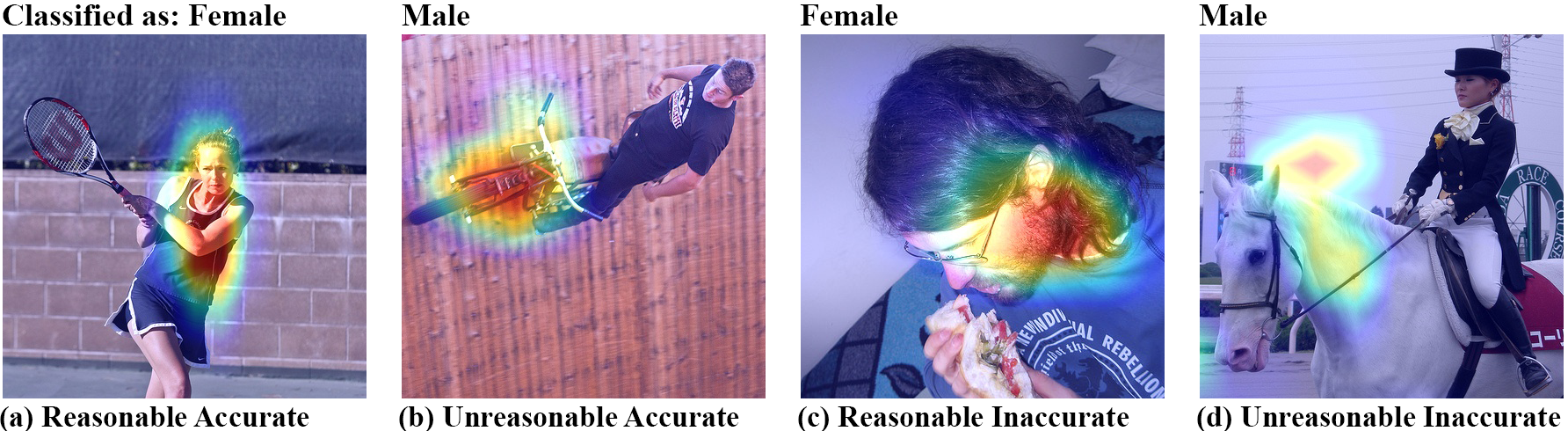}
    \caption{\textbf{Reasonability Matrix} at the top with the four examples in a gender classification problem: (a) \textbf{Reasonable Accurate}: the attention given to an image is reasonable while prediction is also accurate, (b) \textbf{Unreasonable Accurate}: a substantial amount of attention is given to ``contextual'' features which make the attention unreasonable while the prediction is accurate, (c) \textbf{Reasonable Inaccurate}: despite the reasonable attention given to gender-intrinsic features, the prediction is not accurate, and (d) \textbf{Unreasonable Inaccurate}: the attention is unreasonable and the prediction not accurate.}
   \label{fig:teaser}   
\end{figure*}

Although the notion of ``bias'' can be broadly defined~\cite{saxena2019fairness, chouldechova2018frontiers}, studies generally agree that one of the major sources that cause bias is associated with the bias already encoded in data used for training models~\cite{hendricks2018women, chouldechova2018frontiers, baeza2018bias}.
In an image classification task, for example, using a training set where the class distribution is highly skewed towards a particular class~\cite{hendricks2018women, chouldechova2018frontiers, barlas2021see} can lead to decreased prediction accuracy for a minority class~\cite{weiss2007cost, kim2019learning}.
Even when using techniques for handling such imbalanced training-set problem~\cite{chawla2002smote, he2009learning}, DNNs can still be vulnerable to contextual bias~\cite{wang2019balanced} which arises when images in the training set have some contextual objects commonly co-occurring with a particular class~\cite{hendricks2018women}.
For example, in a gender classification problem, model explanation results can give attention to contextual objects, such as baseball bats, snowboards, or kitchenware, rather than intrinsic features directly related to classifying a gender~\cite{wang2019balanced}.
The negative consequences caused by such biased model attention is well represented in recent studies through their catchphrase, such as Hendricks et al.'s ``women also snowboard''~\cite{hendricks2018women}, or Zhao et al.'s ``men also like shopping''~\cite{zhao2017men}.
Steering DNNs to a direction that decouples ``spurious correlation'' has been an important and open research topic~\cite{sagawa2020investigation}. 

Our observation is that using computational approach to model and detect cases of spurious correlation can be complex and challenging ~\cite{quadrianto2019discovering, hendricks2018women} while ``just noticeable'' using human eyes.
Based on this observation, we design, devise, and evaluate the framework of \frameworkfull (\framework, hereinafter) that implements feedback loop between humans and DNNs.
\framework introduces the way for humans to give a case-based feedback to DNNs using explanation methods (or widely known as model attention~\cite{selvaraju2017grad}).
For instance, heatmaps overlayed on images in Fig.~\ref{fig:teaser} display model attention for a gender classification problem generated from the gradient saliency map.
Model attention given in (b) on bicycles and (d) on a horse can be the results of a contextual bias. 
In that case, human annotators can adjust the biased attention through an annotation interface.
Using the human-adjusted attention maps, our approach update the model behavior.
Our \textit{methodological framework} of \frameworkfull presents two novel approaches in the following two stages: In the first stage, in selecting what cases to adjust, we propose \textit{\textbf{reasonability matrix}}, which  uses \textit{attention accuracy} that represents whether the attention given to an image is reasonable from human eyes, on the top of the instance is classified accurately. 
Using two accuracy types, the matrix categorizes every instance into four types (see the top table in Fig. ~\ref{fig:teaser}).
In the second stage of our framework, we present \textit{\textbf{\technique}}, a novel DNN fine-tuning pipeline that jointly maximizes the accuracy and reasonability of the predictions via back-propagation on both prediction and attention loss based on the extension of Grad-CAM~\cite{selvaraju2017grad}.
While a few former work, such as Attention Branch Network (ABN)~\cite{mitsuhara2019embedding} demonstrates the technical feasibility of applying a human's adjusted attention in improving DNNs, there has been no approach that attempt to handle contextual bias through systemic detection, adjustment, and update.


We conducted two studies to evaluate our methodological framework of \framework and a fine-tuning technique, \technique.
In our evaluation, we used 5,000 images collected from Microsoft COCO dataset~\cite{lin2014microsoft} that half show females and the rest show males.
In Study 1 (S1), we considered the following conditions in selecting instances for fine-tuning a DNN model: (1) first baseline only using inaccurate instances and also not using attention loss when fine-tune, (2) the second baseline using inaccurate instances but applying both prediction and attention loss.
For our first experimental condition, we now chose (3) unreasonable instances that humans picked and applied \technique. Finally, in (4) second experimental condition, we used both inaccurate and unreasonable instances using \technique. The results show that the quality of attention measured computationally and human-assessment has been increased significantly from the order of (1), (2), (3), and (4). This indicates the effect of applying the Reasonability Matrix in steering DNNs.

In Study 2 (S2), we tested how \technique can improve ABN, the state-of-the-art approaches for using human-adjusted attention, in two directions: understanding the improvement for attention quality (S2-1), and model performance (S2-2).
In S2-1, we found the quality of attention made based on \technique significantly outperformed than ABN-based approach. While both ABN and \technique all tended to produce reasonable attention maps than the approach not using the adjusted maps, we also found \technique made the attention more effectively exclude contextual objects in general.
In S2-2, we found the predictive power of DNNs can be also improved by applying the \technique in the scenarios where the training samples are limited. While we observed that both \technique and ABN-based approach can leverage the additional attention labels to improve baseline model performance, \technique is much more effective and can achieve a significantly larger improvement than ABN-based approach under different few sample training scenarios studied.
 
This work offers the following contributions:
\begin{itemize}
  \item \textbf{Methodological contribution}: we formalize a methodological framework of \frameworkfull which can effectively debias DNNs' prediction patterns through human's model attention adjustment.
 \item \textbf{Technical contribution}: we present \technique, a novel computational pipeline that jointly maximizes the accuracy and reasonability of predictions based on back-propagation that strikes the balance between prediction accuracy and attention accuracy in fine-tuning DNNs.
 \item \textbf{Empirical contribution}: we present the results of the two studies that indicate the effect of \framework and \technique in steering DNNs.
\end{itemize}
\section{Related Work}

Debiasing Machine Learning (ML) models starts from understanding how the models behave. Therefore, we introduce how ML communities and HCI attempt to connect humans and models through a human-in-the-loop approach to improve model quality. Then we cover computational approaches for debiasing models in ML communities.

\subsection{Human-in-the-loop Approaches for Connecting Humans and DNNs}
Human-in-the-loop has been a promising domain in building superior DNNs by presenting more direct ways to inject human perspective into DNNs. We categorize former approaches into two: principle-driven and annotation-driven.

Principle-driven approaches directly embedding predefined human-driven principles to a model, such as distribution and attribution priors, as inductive bias into the models. For instance, logic rules or rationales are augmented to the training process of DNNs~\cite{hu2016harnessing, zhang2016rationale}. This type of human knowledge embedding is typically specific to and tailored for only special model types and application domains. 
Annotation-driven can be a useful strategy when principles or rules are not feasible to be embedded. One relevant work is Attention Branch Network (ABN)~\cite{fukui2019attention}, a technique that enables adjustment of a model's attention given to an image~\cite{mitsuhara2019embedding}. Another approach within this category uses feature attribution methods to human priors in building DNN~\cite{liu2019incorporating}.
This new type of human annotation approach presents a way to regularize some undesirable samples toward correcting attention~\cite{mitsuhara2019embedding} and salience patterns~\cite{liu2019incorporating} to enhance the reasonableness of the model prediction process. However, applying ABN is not without limitations. For instance, it requires modification of base model architecture which limits its extensibility. Also, the attention maps in ABN are directly generated by a separate network branch which increases the risk of overfitting on the human-adjusted attention maps. 

Annotation-based approaches have been investigated recently, and some studies started applying the approach for different data types, such as images~\cite{mitsuhara2019embedding, shao2020towards}, texts~\cite{jacovi2020aligning, ross2017right}, and attributed data~\cite{visotsky2019few}. 
The approaches demonstrate useful insights into building interactive tools between humans and DNNs, While there has been a lack of research from the human factor side in terms of how we can successfully elicit human knowledge and how such elicitation can be measured from the human's mind. Meanwhile, the former studies are generally not model-agnostic which confine its extensibility. Finally, while the body of de-biasing of DNN literature is growing, we found relatively few approaches in human-in-the-loop (e.g., feature attribution method~\cite{liu2019incorporating}).


\subsection{Interactive Tools for Bridging Humans and ML}
Embedding human knowledge in steering DNNs starts from humans to understand the way DNNs work. In understanding DNN's behavior, it's worth understanding how interactive tools and visual analytics evolved to support humans working on building, understanding, and debugging DNNs~\cite{Kahng2017activis, pezzotti2017deepeyes}.

Ever since the early release of Weka~\cite{hall2009weka} and Fail et. al's notion of Interactive Machine Learning~\cite{fails2003interactive}, communities in HCI and Visualization have designed, built, and deployed interactive systems to improve the interaction between humans and models. EnsembleMatrix presents a way to lineally combine multiple models to create a new~\cite{talbot2009ensemblematrix}. \textit{iVisClassifier}~\cite{choo2010ivisclassifier}, Alshallakh et. al's system~\cite{alsallakh2014visual}, \textit{Squares}~\cite{ren2016squares}, and \textit{RegressionExplorer}~\cite{dingen2019regressionexplorer} applied visual analytic approaches for investigating ML models. FeatureInsight presented a design for ideating features in building a model~\cite{brooks2015featureinsight}, while \textit{ModelTracker} proposed a system that support a life cycle of model building process through interactive visualization~\cite{amershi2015modeltracker}. \textit{MLCube Explorer} presents a way to compare multiple models~\cite{kahng2016visual}. As ML model's types and the way to build introduce the trade-offs between transparency and model performance, some later work, such as \textit{Prospector}~\cite{krause2016interacting}, \textit{RuleMatrix}~\cite{ming2018rulematrix}, and \textit{VINE}~\cite{britton2019vine} suggest interaction techniques or systems that can increase transparency of black-box-based models' prediction patterns and cases.

With the advent of DNNs, another line of research focused on understanding DNN's behavior with the power of interactive tools~\cite{Kahng2017activis, pezzotti2017deepeyes}. LSTMVis ~\cite{strobelt2017lstmvis} and RNNVis~\cite{ming2017understanding} proposed visual analytic systems for RNNs, improving the way people perform sequence modeling. Liu et. al~\cite{liu2016towards} and Bilal et al.~\cite{bilal2017convolutional} presented a visual analytics tool dedicated to examining CNNs. Finally, Liu et. al~\cite{liu2017analyzing} and GAN Lab~\cite{kahng2018gan} examine the visual analytic approaches to better understand the process for building generative models. One of the notable other approaches includes the TensorFlow Graph Visualizer that helps advanced users to understand the complex architecture of DNNs~\cite{wongsuphasawat2017visualizing}.

The approaches present a rich presentation of model behavior through visualization and interaction. However, relatively little work presents a modality that humans can leverage to directly steer the way DNNs operate. Also, the community is in the early stage of exploring fairness and discrimination issues in ML; a few notable approaches, such as FairVis~\cite{cabrera2019fairvis} and Silva~\cite{yan2020silva} have been proposed.

\subsection{Computational Approaches for De-biasing ML}
Bias and discrimination are critical issues in ML systems~\cite{chouldechova2018frontiers}, and a fast-increasing number of approaches have been proposed in ML communities to tackle inherent bias present in data as well as the complex interaction between data and learning algorithms~\cite{zliobaite2015survey, pedreshi2008discrimination}.
Existing approaches typically can be categorized into pre-processing-based, in-processing-based, and post-processing-based approaches.

Pre-processing approaches adjust the data distribution to guarantee a ``fair'' representation of the different classes in the training set~\cite{kim2019learning, quadrianto2019discovering}. The assumption is that if a model is trained on discrimination-free and ``balanced'' data, its predictions will not be discriminatory~\cite{kamiran2009classifying, calders2009building}. Despite the effectiveness of such approaches, however, more recent studies show evidence that the pre-processing-based method may not warrant models without bias~\cite{aghaei2019learning}. 
One notable challenge is \textit{intrinsic biases} that arise in a data-balanced setting due to latent, unlabeled, and ``spurious'' correlation between contextual features (e.g., kitchenware such as dishes) closely associated with a particular class (e.g., female class)~\cite{wang2019balanced}.
In-processing methods improve existing learning algorithms to account for fairness as well, instead of merely the predictive performance~\cite{calders2010three, hardt2016equality}. Different from model-agnostic pre-processing, methods that fall into this type are algorithm-aware, although methods devised for DNNs are rather scarce.
Post-processing-based approaches adjust the resulting models by ``correcting'' the decision boundaries that lead to redlining for a fair representation of different subgroups in the final decision process~\cite{kamiran2010discrimination}.
Although there is a breadth of approaches discussed in this area, this line of research focuses relatively less on directly eliciting human knowledge in refining ML models based on the weaknesses observed in a validation set~\cite{nushi2018towards}.

Our review reveals the potential of the annotation-based human-in-the-loop approach as a way to more directly embed human knowledge in steering DNNs, while dedicated human-factor-centered research in this direction is missing and improvement on the computational side are required. Meanwhile, we identified intrinsic bias caused by the latent co-occurrence of target objects and contextual objects in a testing set as an open research problem. Finally, such problem space can be a suitable testing bed to understand the effect of steerable DNNs.
\section{Methodological Framework of Interactive Attention Alignment}

\begin{figure*}[!t]
\includegraphics[width=\textwidth]{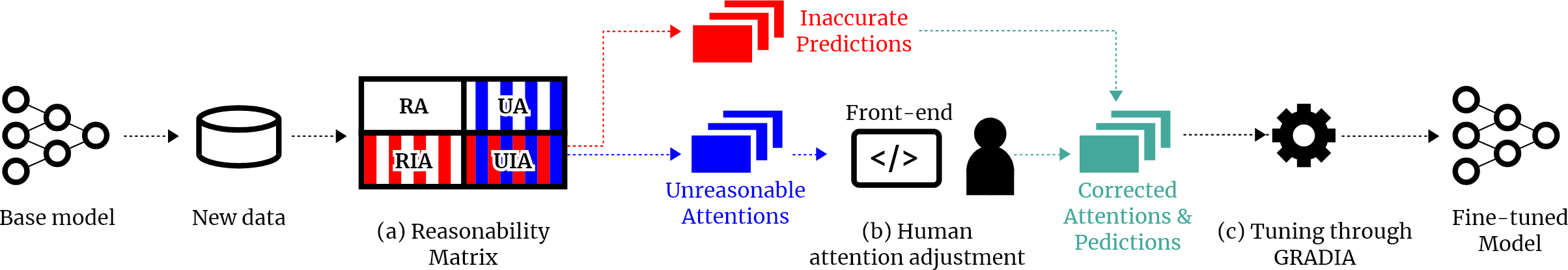}
\caption{Overview of our methodological framework of interactive attention alignment. (a) Building Reasonability Matrix, (b) adjusting attention maps of inaccurate predictions \& unreasonable instances, (c) fine-tune the model using \technique.}
\label{fig:process}
\vspace{-10pt}
\end{figure*}

We elaborate on our methodological framework of \framework devised for steering the way DNNs ``think'' based on human knowledge. Our framework has two notable components: (1) What to adjust: building of the Reasonability Matrix for systemically detect instances where the attention is unreasonable/biased and requires adjustment, and (2) How to adjust: applying \technique to leverage the adjusted attention maps in improving DNNs. Our framework is depicted in Fig.~\ref{fig:process}.

\subsection{What to Adjust: Reasonability Matrix}

The first stage the framework suggests is to identify the parts that display biased attention. Based on the notable work in the literature of fairness~\cite{hendricks2018women, wang2019balanced}, we propose Reasonability Matrix as a way to diagnose the current status of DNNs (a table in Fig.~\ref{fig:teaser}).

On top of prediction accuracy, the matrix combines a new dimension of how \textit{reasonable} it is for a model to classify an image into a particular class based on the attention given to an instance. Specifically, we postulate that a human annotator can determine the whole, or some part of, attention given to an image is either \textit{intrinsic attention}--the attention directly relevant for a classification--or \textit{contextual attention}--the attention that shows ``spurious correlation'' between the object and a specific class (e.g., kitchenware and female, or a baseball hat and male).
To help annotators to decide as to whether attention given to an instance is \textit{reasonable}, we propose the following two-step validation. First, examining intrinsic attention: is the attention given to an image presents sufficient details for a human annotator to classify the instance? If the outcome is positive, the human annotator goes to the next step, examining contextual attention. Using the attention given to an image, can a human annotator recognize any contextual objects? The human annotator can determine the attention is reasonable only if (s)he were not able to recognize contextual attention. To the end, the capability of validating attention reasonability leads to the four cases
as follows:
\begin{itemize}
    \item \textbf{Reasonable Accurate (RA)}: The attention given in an instance is focusing on intrinsic features directly relevant to a classification task while a model's prediction is accurate. Screening out the areas without attention would give a human annotator sufficient evidence to classify the instance.
    \item \textbf{Unreasonable Accurate (UA)}: The prediction itself is accurate, but a non-trivial amount of attention is given to contextual features. This type of attention may indicate the intrinsic bias~\cite{wang2019balanced} which needs to be adjusted (Fig.~\ref{fig:teaser} (b), the attention is given to a bicycle rather than a man).
    \item \textbf{Reasonable Inaccurate (RIA)}: Even though the attention is focusing on intrinsic features, the outcome is inaccurate. This might be due to the lack of data points in a train set similar to this type (Fig.~\ref{fig:teaser} (c), attention is given to a man's beard but the model makes an inaccurate prediction).
    \item \textbf{Unreasonable Inaccurate (UIA)}: The attention is not reasonable and also the prediction is inaccurate.
\end{itemize}

Recent studies report the raising concerns in the data science community about their practice of heavily relying on a single error score, accuracy performance metric, or confusion matrix in evaluating ML models. 
It is the general consensus that those methods may not present a comprehensive picture in capturing a model's ``crucial shortcomings''~\cite{nushi2018towards}.
The capability of structuring the reasonability matrix implies we can use the quality of attention to evaluate DNN's performance in new perspectives. On top of the widely used model prediction accuracy metric, our framework proposes the following metrics as new ways to add more rigor in evaluating DNN:

\begin{itemize}
    \item \textbf{Reasonable Accurate Performance}: the metric indicates a more rigorous model performance than generally used prediction accuracy performance derived by measuring the proportion of RA among every instance.
    \item \textbf{Attention Accuracy Performance}: the metric explains the quality of model attention by measuring how many instances that DNN gave reasonable attention (i.e., the sum of RA and RIA) among every instance. This metric can be a direct proxy that how models' attention focuses on parts directly relevant for the prediction which can mean the impact of intrinsic bias.
\end{itemize}

\begin{figure}[bp]
\centering
\includegraphics[width=0.4\textwidth]{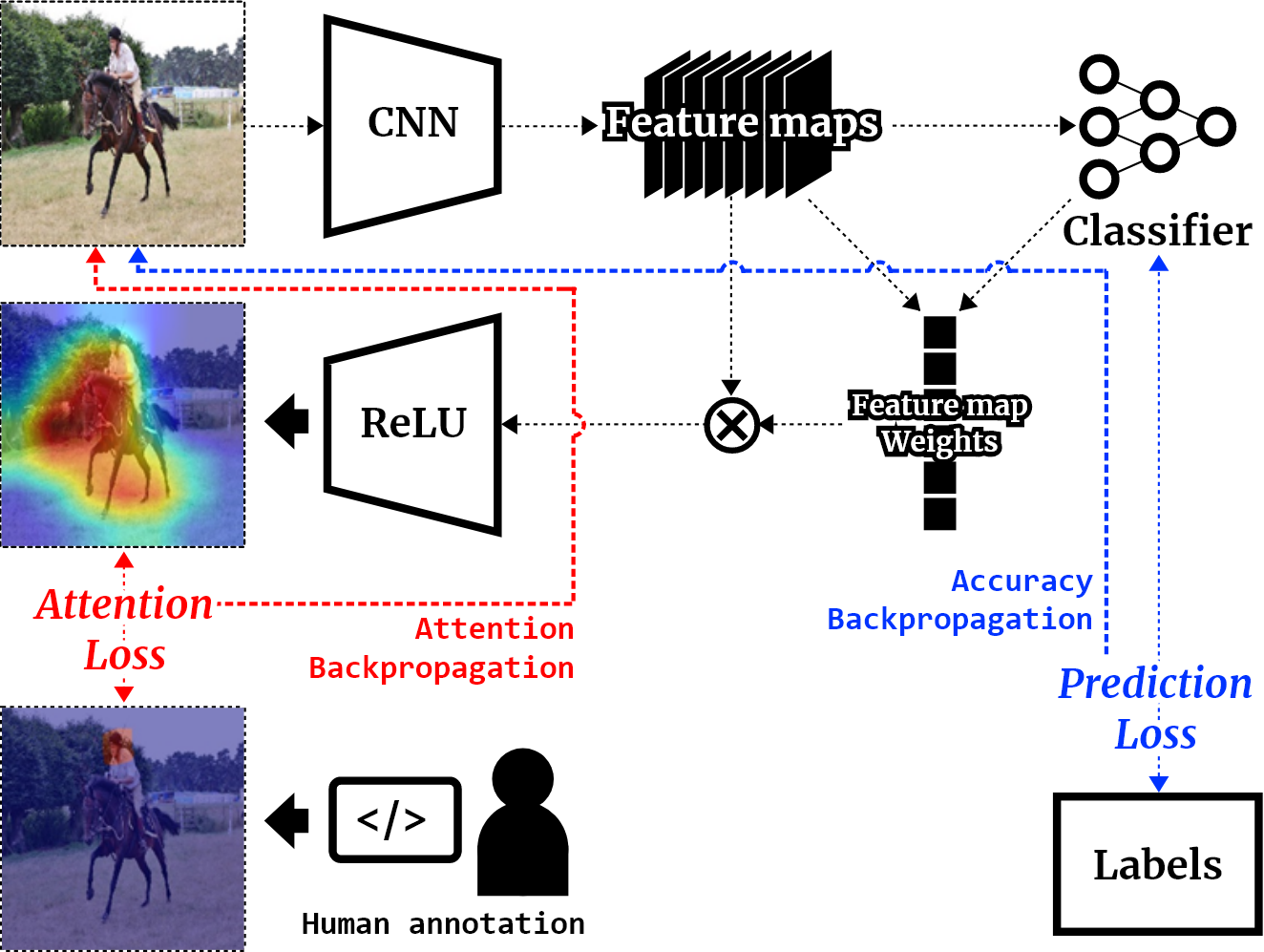}
\caption{The computational pipeline of \technique.}
\label{fig:architecture}
\end{figure}

\subsection{How to adjust: \technique}

Using the proposed Reasonability Matrix, our framework elicits adjusted attention from human annotators. In this section, we introduce how \technique uses the adjusted attention maps in fine-tuning DNNs.


In addition to minimize the error in the original training set, our major goal is to also minimize the losses from the three terms UA, RIA, and UIA in the Reasonability Matrix, which directly leads to our objective:
\begin{equation}
\label{eq:loss1}
\min \ \mathcal{L}_{\text{Train}} + \mathcal{L}_{\text{UA}} + \mathcal{L}_{\text{UIA}} +  \mathcal{L}_{\text{RIA}}
\end{equation}
where $\mathcal{L}_{\text{Train}}$ denotes the model prediction loss on the original training set; $\mathcal{L}_{\text{UA}}$, $\mathcal{L}_{\text{UIA}}$, and $\mathcal{L}_{\text{RIA}}$ measure the errors on Unreasonable Accurate (UA), Unreasonable Inaccurate (UIA), and Reasonable Inaccurate (RIA) samples according to the quadrants in the Reasonability Matrix of validation set, respectively. 



For each term in Equation \eqref{eq:loss1}, there are two types of losses, namely prediction loss, denoted by $\mathcal{L}^{(p)}$, and attention  loss, denoted by $\mathcal{L}^{(a)}$. Considering that different term (from different quadrant in Reasonablity Matrix) requires different focus and balance between prediction and attention, we further introduce the balance factors for each term to give the model the flexibility to better weight between the attention and prediction loss in different cases.
Specifically, Equation \eqref{eq:loss1} can be expanded into the following one:
\begin{equation}
\label{eq:loss_2}
\min \  
\mathcal{L}_{\text{Train}} +  
(\alpha \mathcal{L}^{(p)}_{\text{UA}} + (1-\alpha)\mathcal{L}^{(a)}_{\text{UA}}) +
(\beta \mathcal{L}^{(p)}_{\text{UIA}} + (1-\beta)\mathcal{L}^{(a)}_{\text{UIA}}) + 
(\gamma \mathcal{L}^{(p)}_{\text{RIA}} + (1-\gamma)\mathcal{L}^{(a)}_{\text{RIA}})
\end{equation}
where the parameters $\alpha$, $\beta$, and $\gamma \in [0,1]$ are the tunable factors for controlling the balance between the prediction loss and attention loss for UA, UIA, and RIA samples, respectively.

This way, the first term $\mathcal{L}_{\text{Train}}$ can also be expanded as a special case $\mathcal{L}_{\text{Train}}=\mathcal{L}^{(p)}_{\text{Train}}$ where the weight for $\mathcal{L}^{(p)}$ is set to 1 and the weight for $\mathcal{L}^{(a)}$ is set to 0, such that the attention map labels are not required.
Finally, by further expanding the first term and rearranging the terms for prediction losses and attention losses, the final objective of \technique can be written as: 
\begin{equation}
\label{eq:loss3}
\mathcal{L}_{\text{\technique}} = \underbrace{
\mathcal{L}^{(p)}_{\text{Train}} + 
\alpha \mathcal{L}^{(p)}_{\text{UA}} + 
\beta \mathcal{L}^{(p)}_{\text{UIA}} + 
\gamma \mathcal{L}^{(p)}_{\text{RIA}}
}_{\mbox{\small prediction loss}}
+
\underbrace{
(1-\alpha)\mathcal{L}^{(a)}_{\text{UA}} +
(1-\beta)\mathcal{L}^{(a)}_{\text{UIA}} +
(1-\gamma)\mathcal{L}^{(a)}_{\text{RIA}}
}_{\mbox{\small attention loss}}
\end{equation}
where $\mathcal{L}^{(p)}$ can be calculated by applying the Cross-entropy loss on the corresponding samples of each terms; and $\mathcal{L}^{(a)}$ is the newly proposed attention loss that measure the attention quality of the samples.

By introducing $\mathcal{L}^{(a)}$ into the fine-tuning step with \technique, the base DNN model can be jointly optimized both to generate higher quality attention maps and to make better and unbiased predictions on the original task. 
Our assumption is that this attention de-biasing process will also enhance the generalizability of the model to unseen data. 
As a result, \technique will ultimately not only improve the model prediction accuracy, but also yield a more interpretable model.

To quantify the attention quality of the model, we propose a general attention loss for estimating the discrepancy between the model-generated attention maps and the human-annotated attention labels of the selected samples from the validation set. Concretely, the attention loss can be computed as the following:
\begin{equation}
\label{eq:attention_loss}
\mathcal{L}^{(a)} = \text{dist}(M, M^\prime)
\end{equation}
where $M$ and $M^\prime$ are the model-generated attention maps and the ground truth attention maps provide by the human annotators on those samples that require attention adjustment; the function $\text{dist}(x, y)$ can be a common divergence metric such as absolute difference or square difference. In practice, we found that absolute difference is more robust to the labeling noise from the annotator, while square difference can be more sensitive and yield a high loss on the border areas of the labels that could not actually be related to the object.

To generate the model attention maps on images, several existing works have been proposed. Response-based methods such as CAM~\cite{zhou2016learning} and ABN~\cite{fukui2019attention} typically require substantial modification on the DNN architectures that either hurt the model's performance and extensibility or over-decouple the generation process of attention and prediction. For example, to handle the performance issue, ABN proposed to add another module called `attention branch' onto the model architecture that is specialized for generating the attention maps. However, this incurs much more parameters and hence more samples and time to train the model. Moreover, over-decoupling the components for producing attention and prediction substantially decreases the reliability that the attention is indeed the explanation for the prediction.
In contrast, gradient-based methods such as Grad-CAM~\cite{selvaraju2017grad} does not require changes of the base model and hence is applicable to a wide range of various DNN models.
Moreover, it does not incur additional model parameters and hence can be more computationally cheap. Furthermore, its attention and prediction are tightly coupled and hence maintain a strong dependency and reliability between the prediction and its attention map.

Therefore, we propose to build our pipeline by extending Grad-CAM which uses the gradient of the feature maps with respect to the target class to generate the attention maps.
Mathematically, suppose the penultimate layer produces $K$ feature maps, $A^k \in\mathbb{R}^{u \times v}$ where $u$ and $k$ are the width and height of the image of each feature map.
The attention maps $M_{\text{Grad-CAM}} \in\mathbb{R}^{u \times v}$ for target class $c$ can be computed as:
\begin{equation}
\label{eq:grad_cam}
M_{\text{Grad-CAM}} = \text{ReLU}(\frac{1}{uv} \sum_k\sum_i \sum_j \frac{\partial
Y^c}{\partial A^k_{i,j}} \cdot A^k)
\end{equation}
where $Y^c$ denotes the output of the model for predicting class $c$, and $\sum_i \sum_j\partial Y^c/\partial A^k_{i,j}$ denotes the weight of the feature map $k$ for class $c$ as also illustrated by Figure \ref{fig:architecture}.
To ensure the generated and labeled attention maps are in the same scale, we further normalize $M_{\text{Grad-CAM}}$ to the values between 0 and 1, as:
\begin{equation}
\label{eq:gc_norm}
M = \frac{M_{\text{Grad-CAM}} - \text{min}(M_{\text{Grad-CAM}})}{ \text{max}(M_{\text{Grad-CAM}}) - \text{min}(M_{\text{Grad-CAM}})}
\end{equation}
where the function $min(\cdot)$ and $max(\cdot)$ return the element-wise min and max  of the input, respectively.

Notice that there are two major differences that distinguish the proposed \technique from the previous works, such as in~\cite{mitsuhara2019embedding} and~\cite{liu2019incorporating}. 
First, instead of only correct the model attention on the misclassified instances, \technique best leverages the Reasonability Matrix to identify which sample's attention needs to be adjusted based on both the prediction accuracy as well as attention accuracy.
Second, \technique offers the flexibility to control the balance between attention accuracy and prediction accuracy for different instances from different quadrants, which enables us to easily find the 'sweet spot' of the model that can produce more reasonable attention without scarifying the model accuracy.
For example, for the instances in quadrant 'Unreasonable Accurate (UA)' of the Reasonability Matrix, we can set $\alpha$ in the range of $(0, 0.5)$ to force the model to pay less attention to prediction loss while paying more attention to the attention loss, as the prediction was accurate but the attention was biased.

In all, we introduced the two core components that establish our framework of \framework. The first component is the Reasonability Matrix for identifying the instances with biased attention that need to be adjusted. Building the Reasonability Matrix requires humans to scan the quality of attention of instances and adjust those when required. This process requires additional human-based computation costs. In Study 1, we investigate if using the Reasonability Matrix in fine-tuning DNNs can counterbalance such costs by improving the model's quality in both computational and human-assessment-based ways.  Meanwhile, the second component is \technique which is devised for optimally utilizing the human-adjusted attention maps to strike the balance between prediction accuracy and attention accuracy in fine-tuning DNNs. To understand the effect of \technique in fine-tuning DNNs, we compare \technique to the state-of-the-art, ABN~\cite{mitsuhara2019embedding} regarding a variety of metrics in Study 2.
\section{S1: Effect of \frameworkfull}
\begin{figure*}[!b]
\includegraphics[width=\textwidth]{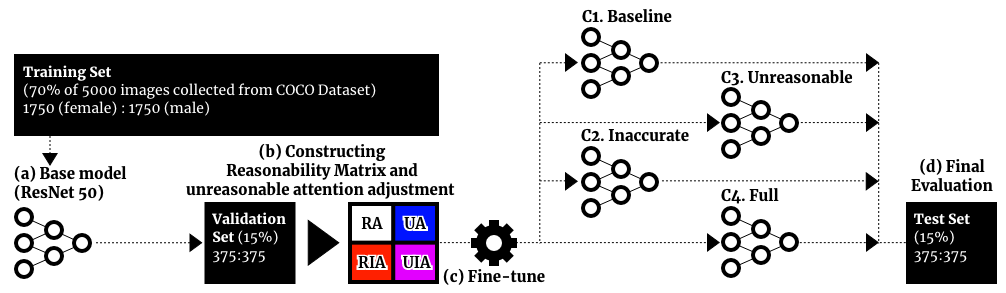}
\caption{Study 1 Methodological process: (a) Using a training set to build a base model, (b) using a base model to build a reasonability matrix using a validation set, (c) fine tune four different model that each represents a study condition, (d) use each model to validate using a test set.}
\label{fig:S1_method}
\end{figure*}

Study 1 (S1) aims at understanding the effect of using Reasonability Matrix in fine-tuning DNNs. In our evaluation, we compared four conditions that include two baselines (conditions that fine-tuned a model without Reasonability Matrix) and two experimental conditions (conditions that used Reasonability Matrix). Our overall process is explained in Fig.~\ref{fig:S1_method}.

\subsection{Methodology}

\subsubsection{Task \& Dataset:} The gender classification problem is one of the widely used tasks in the research of fairness in ML~\cite{zhao2017men, barlas2021see}. 
We are aware that using a binary classification in gender does not reflect on the diverse viewpoint of gender in the real world, as the term ``gender'' itself includes subtle social roles and personal identification and the outcome can be more diverse than binary. 
We chose gender classification in a binary scheme because it is easy to understand the effect of \framework than more complex settings such as multi-class classification.
Also, the Microsoft COCO dataset~\cite{lin2014microsoft} we use has a highly well-annotated binary gender label so that is a good resource for the binary study.
We emphasize that the binary ``gender classification'' 
task established in broader ML communities ~\cite{zhao2017men, barlas2021see, hendricks2018women} does not represent our viewpoint on gender.


For the study, we constructed our dataset from the Microsoft COCO dataset~\cite{lin2014microsoft}, which has been widely used in ML research~\cite{zhao2017men}. To build our dataset, we extracted images that had the word ``men'' or ``women'' in their captions. We then filtered out instances that (1) contain both words, (2) include more than two people, or (3) humans appear in the figure is nearly not recognizable from human eyes. For the scope of our problem, we manually selected images until we reach to collecting 2,500 female and 2,500 male images. In our settings, we used the data split of 70\%:15\%:15\% for training set, validation set, and test set (see black boxes in Fig.~\ref{fig:S1_method}.)

\subsubsection{Conditions}
Using the the training set (3,500 images), we built a baseline model based on the ResNet50 architecture~\cite{he2016deep} (see Fig.~\ref{fig:S1_method} (a)). We used the baseline model in validating images in the validation set. Specifically, we asked a human annotator to assess the quality of attention and built the Reasonability Matrix (see Fig.~\ref{fig:S1_method} (b)). In eliciting their assessment, the annotator had to answer the two reasonability questionnaires as follows: \textbf{Q1.} Does the focus area contains necessary details enable you to classify a gender? (Yes/No) \textbf{Q2.} Does the focus area contains unnecessary details not related for you to classify a gender? (Yes/No)
Only the instances that the annotator answered ``Yes'' in Q1 and ``No'' in Q2 were considered reasonable. Among 750 instances, the annotator answered 232 unreasonable. For the instances assessed either as unreasonable or inaccurate, the annotator used a line drawing interface we provided to draw a binary mask that shows the areas (s)he felt the attention should be given to be reasonable. Using the Reasonability Matrix and the adjusted binary attention maps that the annotator provided, we fine-tune the four different models that corresponding to four conditions (Fig.~\ref{fig:S1_method} (c)):
\begin{itemize}
    \item \textbf{C1. Baseline}: We fine-tuned the model only using the prediction loss; no human labeled attention maps were used.
    \item \textbf{C2. Inaccurate}: In this condition, we fine-tuned the model via the prediction loss and a subset of the attention loss from inaccurate instances (i.e. RIA and UIA quadrants), which in total used 119 attention map labels. Notice that C1 and C2 does not necessarily require building Reasonability Matrix, as one can identify what to adjust based solely on the prediction without assessing the reasonability.
    \item \textbf{C3. Unreasonable}: This condition applies the prediction loss and a subset of attention loss from unreasonable instances (i.e. UA and UIA quadrants), which in total used 232 attention map labels.
    \item \textbf{C4. Full}: This condition fully utilizes both prediction and attention loss from instances that are inaccurate and/or unreasonable (i.e. UA, RIA, and UIA quadrants) as shown in Equation \eqref{eq:loss3}, which in total used 295 attention map labels. C3 and C4 requires establishing Reasonability Matrix as every unreasonable instances will be considered in the attention loss for fine-tuning the model .
\end{itemize}

\subsubsection{Measures \& Study Apparatus}
\begin{figure*}[!t]
    \includegraphics[width=\textwidth]{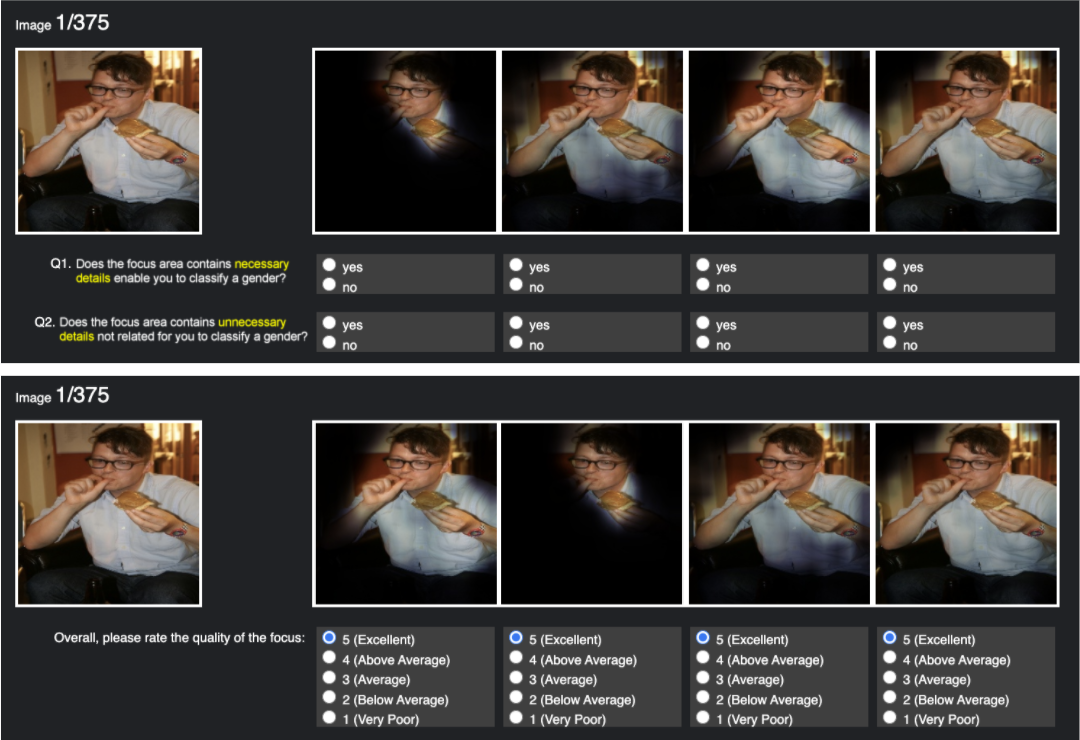}
    \caption{Study apparatus used for eliciting M4. Attention Accuracy (top), and M5. perceived attention quality (bottom).}
    \label{fig:S1_Apparatus}
\end{figure*}

We evaluate the quality of the four models using the test set (see Fig.~\ref{fig:S1_method} (d)). In particular, we deployed the following measures to quantitatively/qualitatively evaluate the performance of the models.
\begin{itemize}
  \item \textbf{M1. Prediction Accuracy Performance}: a measure that shows a model prediction accuracy.
  \item \textbf{M2. ``RA'' Performance}: a measure that shows the proportion of Reasonable Accurate (RA) out of every instance.
  \item \textbf{M3. Intersection over Union (IoU)}: a measure the quantitatively assess the quality of attention. Following the work on network dissection~\cite{bau2017network}, we collect the ground truth attention form a human annotator for every instance in a test set. Then we make bit-wise intersection and union operations with the ground truth attention maps and each model's attention maps to measure how well the two attention masks overlap.
  \item \textbf{M4. Attention Accuracy Performance}: a measure that shows whether human annotators perceived the attention given to each image was reasonable. For each image in a test set, the annotators saw four different attention that represents the four conditions in a random order. The annotators used the two reasonability questionnaires aforementioned (see Fig.~\ref{fig:S1_Apparatus}, top).
  \item \textbf{M5. Perceived Attention Quality}: a measure that shows the perceived quality of attention. Five-level Likert scale starting from Very Poor (1) to Excellent (5) is used to capture perceived attention quality from human annotators. Like M4, four attention visualizations are presented in a random order (see Fig.~\ref{fig:S1_Apparatus}, bottom).
  \item \textbf{M6. Post-hoc Result Analysis}: Finally, we analyze the resulted attentions across the four conditions with our eyes to identify the patterns and tendency.
 \end{itemize}
 
\subsubsection{Human-based Assessment}
M4 and M5 required human-based assessment; each human annotator should see the four different attention results of 750 images and compare the attention quality. To acquire assessment results ran based on a consistent standard, we aimed at asking each annotator to assess the whole instances. If we assume each annotator uses 10 seconds to assess one attention, assessing the whole attention (750 instances x 4 conditions) will take more than 8 hours (without counting the time to refresh their attention). Such multiple-days commitment is not suitable for crowdsourcing. Instead of using crowdsourcing platforms, we recruited 3 participants for M4 and 7 participants for M5 using a public university in the United States through word-of-mouth and on-campus flyers. For M4, we recruited odd number to remove the case that makes tie. For both M4 and M5, our post-hoc power analysis results for understanding the effect size is reported in Results section.

Upon the agreement of participating in the study, we sent a 15-minutes video that explain the concept of (1) ``attention'' in DNNs, (2) ``intrinsic bias'', and (3) two criteria of inclusion---whether the attention includes enough information for a human to classify a gender in an image, and exclusion---whether the attention exclude contextual objects that are not directly relevant for a classifying a gender in an image. After they watch the video, we presented a quiz that asks three questions related to attention quality assessment. Once participants complete the quiz, they either used apparatus in Fig.~\ref{fig:S1_Apparatus}, top (for those assigned to M4) or apparatus in Fig.~\ref{fig:S1_Apparatus}, bottom (for those assigned to M5). We advised participants to take enough rest to refresh their attention when mentally exhausted. We also allowed them to label multiple days. On average, participants spent 5 days to complete the assessment. Upon their completion, we compensated our participants with a gift card with \$60 of value.

\subsection{Results}
\begin{table*}[!t]
    \centering
    \small
    \vspace{-3mm}
    \begin{tabular}{>{\raggedright\arraybackslash}p{24mm}|p{14mm}p{18mm}>{\raggedright\arraybackslash}p{13mm}p{13mm}p{13mm}>{\raggedright\arraybackslash}p{13mm}>{\raggedright\arraybackslash}p{13mm}}
        \Xhline{3\arrayrulewidth}
        \textbf{Conditions} & \textbf{Adjusted attention \#} & \textbf{Reasonability Matrix} & \textbf{M1} & \textbf{M2} & \textbf{M3} & \textbf{M4} & \textbf{M5} \\ 
        \Xhline{2\arrayrulewidth}
        \textbf{C1. Baseline} & 0 (0\%)  & $\begin{bmatrix} 306 & 310\\ 33 & 101\end{bmatrix}$ & $82.13\%$ & $40.80\%$ & $0.23 \pm 0.19$ & $45.20\%$ & $2.82 \pm 1.13$  \\
        \textbf{C2. Inaccurate} & 119 (3\%)  &  $\begin{bmatrix} \ 456 & \ 163 \\ \ 87 & \ 44 \end{bmatrix}$ & 82.53\% & 60.80\% & $0.32 \pm 0.19$ & 72.40\% &$3.68\pm1.16$\\
        \textbf{C3. Unreasonable} & 232 (5\%)  &  $\begin{bmatrix} \ 497 & \ 117 \\ \ 99 & \ 37 \end{bmatrix}$ & $81.86\%$ & 66.27\% & $0.34 \pm 0.18$ & 79.47\% &$3.81\pm1.13$\\
        \textbf{C4. Full} & 295 (6\%)  &  $\begin{bmatrix} \ 518 & \ 104 \\ \ 94 & \ 34 \end{bmatrix}$ & \textbf{82.93}\% &  \textbf{69.07}\% & \textbf{0.36$\pm$0.19} & \textbf{81.60\%} & \textbf{3.97$\pm$1.08} \\
        \Xhline{3\arrayrulewidth}
    \end{tabular}
    \caption{S1 Results that show how measures (M1: Prediction accuracy performance, M2: Reasonable Accurate performance, M3: IoU, M4: Attention accuracy performance, and M5 Perceived attention quality) change across the conditions (C1: Baseline, C2: Inaccurate, C3: Unreasonable, and C4. Full. The best performing condition is bolded.}
    \label{table:S1_results}
\end{table*}

The results show that C4 outperformed other conditions in every measure. In terms of M1, C4 demonstrated the best prediction accuracy performance although the gap was not significant. One notable aspect in this observation is that adding more manual input did not hinder the model performance. Comparing to the baseline, C2 and C4 demonstrated slight improvement while C3 showed 0.27\% of performance drop. In their recent study, Barlas et al. shared their observation of ``accuracy-fairness gap'' which denotes the case where embedding human's perspective in steering black-boxes can result in dropping the model accuracy performance~\cite{barlas2021see}. Although our study results didn't show the accuracy-fairness gap, the performance drop/increase may become obvious or significant if using different split ratio of train validation test sets, or use different loss function. 

\begin{figure*}
   \centering
   \includegraphics[width=\textwidth]{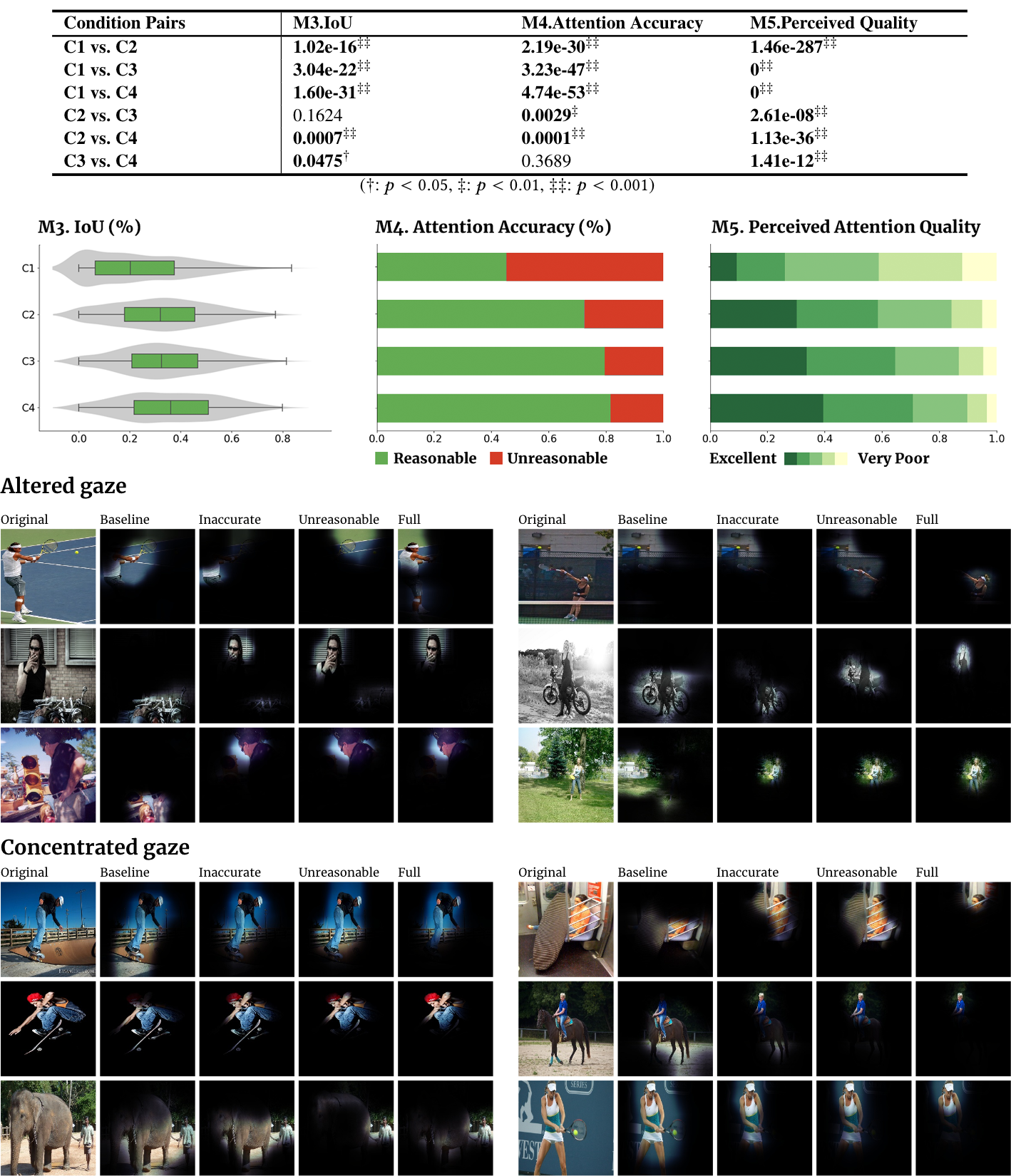}
    \caption{The pairwise comparison results of M3, M4, and M5 (Top), distribution charts (Middle), and attention results (Bottom)}
   \vspace{-5mm}
   \label{fig:S1_pairwise}
\end{figure*}

The rest of the measures are related to the quality of model attention. We found the three measures of M3 (a computationally driven measure), and M4 and M5 (human assessment-based measures) agree that C4 (Full) presented the best attention quality, followed by C3 (Reasonable), C2 (Inaccurate), and C1 (Baseline). The gaps between the four conditions were generally significant. Since the independent variables we collected don't follow the normal distribution, we applied Kruskal-Wallis H-test. 

In terms of M3 of IoU, we found a significant difference within the four conditions (p=$4.13e-34, < 0.05$). Dunn's test for post-hoc pairwise comparison found that every pair was significantly different except the pair of C2 and C3 (see Fig.~\ref{fig:S1_pairwise}, M3 related information). In terms of M4 Attention Accuracy; the majority vote results collected from three labelers in terms of how many reasonable attentions were presented in each condition, Kruskal-Wallis H-test found that the four conditions are significantly different with a p-value of 1.20e-64 ($< 0.05$). Dunn's test found significant difference between every pair except the pair of C3 and C4 (see M4 in Fig.~\ref{fig:S1_pairwise}). Since the approach applied the majority vote, it's important to check the degree to which the three annotators' agreed. Based on the Cohen's Kappa Agreement test, their results showed each pair of annotators ranged from 0.75 to 0.8, which are considered as having ``Substantial agreement'' and ``Almost perfect agreement.'' Finally, M5, the Perceived Attention Quality analyzed 5,250 rating images assessed by 7 participants. With these many observations and the calculated effect sizes (Cohen’s d) for the conditions, the collected data has enough power (over 80\%) for conducting further tests. The p-value yielded using Kruskal-Wallis H-test is close to 0 while Dunn's post-hoc pairwise test found the perceived qualities of all pairs are significantly different (see Fig.~\ref{fig:S1_pairwise}, M5).

For conducting the post-hoc power analysis for the Attention Accuracy Performance, we first calculated the Cohen’s d effect sizes for the condition pairs. All the condition pairs that involve C1 have medium to large effect sizes: C1 vs. C2 (-0.57), C1 vs. C3 (-0.76), and C1 vs. C4 (-0.82). According to the power analysis results, almost all the condition pairs have enough statistical power except the condition pair of C3 and C4, which has a power below 80\%.

By computing the Cohen’s d effect size of the Perceisved Attention Qualities for the condition pairs, we found large effect sizes on three pairs of conditions: C1 vs. C2 (-0.75), C1 vs. C3 (-0.87), and C1 vs. C4 (-1.04). So, when comparing condition C1 with any other condition, a large but negative effect was detected. Even though other condition pairs' effect sizes are relatively small, with the large number of observations collected, the post-hoc power analysis tells us that we have enough statistical power (all above 80\%) to detect any significant difference between the conditions’ qualities, if we set the significance level as 0.05.

In the last part of our analysis, we displayed 4 attention types generated using C1-C4 for every of 750 images. Then we analyzed how the attention varies/changes depending on the condition. As a result, we were able to find two notable patterns. In case the attention given to an instance in C1 does not include sufficient ``intrinsic'' objects (i.e., the attention doesn't give a sufficient clue of sexuality of a person), the gaze tended to shift to the object(s) that are more relevant to classifying gender. Some examples of such ``altered gaze'' is presented in Fig.~\ref{fig:S1_pairwise} where the initial gaze was posed on some objects such as tennis rackets, motorcycles, doll, or background was shifted to humans. The next pattern is the concentration of the gaze. As we browse the attention from C1 to C2, C3, and C4, we found a tendency that the areas-of-focus tend to become smaller, focused, and concentrated. This ``concentrated gaze'' tended to make a positive effect when the baseline gaze includes some contextual objects which would gradually be excluded as the attention is shifted to C2, C3, and C4. Such examples are displayed in Fig.~\ref{fig:S1_pairwise} where the earlier attention includes a human with the contextual object such as skateboard, animals, and surf exclude the contextual objects in C3 and/or C4.  

\subsection{Discussion}
Through the study results, we conclude that the core effect of applying Reasonability Matrix in solving the problem of``what to adjust'' is to increase the quality of attention that aligns with human expectations than what the current tools can offer without sacrificing the model prediction accuracy performance. In fact, numbers of ``Reasonable Accurate'' in Reasonability Matrix in Table ~\ref{table:S1_results} (left top) have increased gradually as we move from C1 to C4. At the same time, we observed that Unreasonables (Unreasonable Accurate and Unreasonable Inaccurate) all gradually decreased. These patterns suggest that (1) instances predicted accurately would more likely because the underlying attention is ``right'' in the later conditions than the former. Also, (2) the inaccurate instances arise because the biased intrinsic attention would likely decrease in the later conditions than former conditions. 

Another interesting pattern we found is the sharp difference of increase of Reasonable Inaccurate between C1 vs C2, C3, and C4. The difference between the two groups is to apply attention loss in fine-tuning the model. We believe this observation gives us the drawback of \technique while show some future opportunity to increase M1 performance for conditions with \technique. As we observed in our qualitative Post-hoc analysis, applying \technique can make the attention more focused. In case the initial attention was posed on the ``right spot'', we often observed that the attention doesn't include anything identifiable from our eyes in C2, C3, and C4 which is an obvious drawback. We assume this resulted in increasing the Reasonable Inaccurate and this factor is one of the types where C2, C3, and C4 will likely make inaccurate prediction. Such observation can give a clue to improve the \technique by penalizing the case where the attention map doesn't include the obvious objects in an image in designing a better objective function for attention loss function.

\section{S2: Effect of \technique}
The goal of S2 is to understand how using \technique in fine-tuning DNNs can improve the quality of DNNs than state-of-the-art. For this benchmark, we chose Attention Branch Network~\cite{mitsuhara2019embedding}. In our evaluation, we chose two directions, one for understanding the degree to which the quality of attention is improved from human eyes (S2-1). Second, understanding if the \technique can improve model performance in the scenarios where the size of sample is scarce (S2-2). 

\begin{figure*}[!t]
\includegraphics[width=\textwidth]{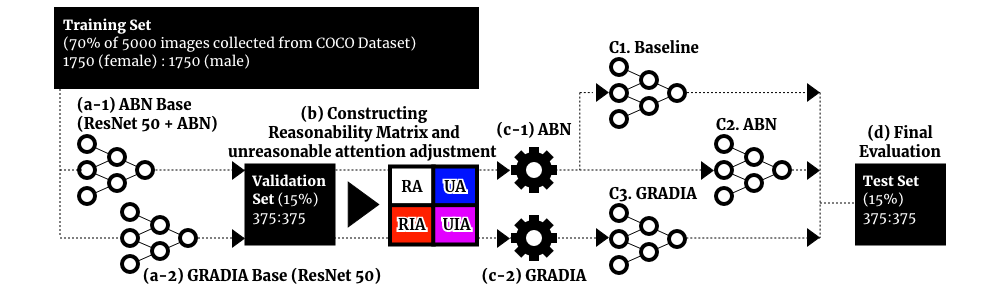}
\caption{Study 2 Methodological process: (a) Using a training set to build two base models, (b) using each base model to construct a reasonability matrix using a validation set, (c) fine-tune three different models that each represents a study condition, (d) evaluate the three models using a test set.}
\label{fig:S2_method}
\end{figure*}

\subsection{S2-1: Assessment through Perceived Quality of Attention}
\subsubsection{Methodology}
In S2, we used the same dataset with the same split, as well as the same task we used in S1. In preparing our conditions, we first used the training set to build two base models; one with ResNet50 + ABN~\cite{fukui2019attention, mitsuhara2019embedding} architecture for building Baseline and ABN later, and another with the base ResNet50 architecture~\cite{he2016deep} for constructing \technique-based approach (see Fig.~\ref{fig:S2_method} (a-1) and (a-2)). We used the two base models in building Reasonability Matrix using the validation set. Similar to S1, a human answered the two reasonability questionnaires. Consequently, we found that 433 instances from the baseline for ABN and 295 from the baseline for \technique need to be adjusted (Fig.~\ref{fig:S2_method} (b)). After identifying all the instances that need to be adjusted, we asked the human labeler to adjust every attention map. Using the adjusted maps, we fine-tuned the three models to build the three conditions as follows:
\begin{itemize}
    \item \textbf{C1. Baseline}: The base ABN model fine-tuned using prediction loss, i.e. without using human adjusted attention (Fig.~\ref{fig:S2_method} (c-1)).
    \item \textbf{C2. ABN}: Beside the prediction loss, we used the additional 433 samples' adjusted attention maps elicited from human labelers to edit the attention of the ABN model~\cite{fukui2019attention} (Fig.~\ref{fig:S2_method} (c-1)).
    \item \textbf{C3. \technique}: The proposed computational pipeline that fully utilizes both prediction and attention loss from instances that are inaccurate and/or unreasonable (i.e. UA, RIA, and UIA quadrants). Note that this condition is the same model we used as C4 Full in S1 (Fig.~\ref{fig:S2_method} (c-2)).
\end{itemize}
Finally, we evaluate the quality of the three models using the test set (see Fig.~\ref{fig:S2_method} (d)). In evaluating the three models, we used the same five measures as well as the study apparatus we used in S1. For eliciting the human assessment, we recruited 3 participants for M4 to perform a majority vote to determine the reasonability of each attention and then 5 participants to evaluate M5 that captures the perceived quality of attention. The participants went through the same procedure of watching 15 minutes of introduction video, performing a short quiz, and eliciting their perspectives using the study apparatus we provided. Upon their completion, we compensated their participation with the same amount with S1.

\subsubsection{Results}

\begin{table*}[!t]
    \centering
    \small
    \vspace{-3mm}
    \begin{tabular}{>{\raggedright\arraybackslash}p{20mm}|p{14mm}p{18mm}>{\raggedright\arraybackslash}p{14mm}p{14mm}p{14mm}>{\raggedright\arraybackslash}p{14mm}>{\raggedright\arraybackslash}p{14mm}}
        \Xhline{3\arrayrulewidth}
        \textbf{Conditions} & \textbf{Adjusted attention \#} & \textbf{Reasonability Matrix} & \textbf{M1} & \textbf{M2} & \textbf{M3} & \textbf{M4} & \textbf{M5} \\ 
        \Xhline{2\arrayrulewidth}
        \textbf{C1. Baseline} & 0 (0\%)  &  $\begin{bmatrix} \ 147 & \ 462 \\ \ 25 & \ 116 \end{bmatrix}$ & 81.20\% & 19.60\% & 0.19$\pm$0.15 & 22.93\% & 2.96$ \pm$ 1.25\\
        \textbf{C2. ABN} & 433 (10\%) & $\begin{bmatrix} \ 38 & \ 580 \\ \ 6 & \ 126 \end{bmatrix}$ & 82.40\%  & 5.07\% & 0.29$\pm$0.18 & 5.87\% & 2.81$ \pm$ 1.16\\
        \textbf{C3. GRADIA} & 295 (6\%)  &  $\begin{bmatrix} \ 515 & \ 108 \\ \ 97 & \ 30 \end{bmatrix}$ & \textbf{82.93}\% &  \textbf{68.67}\% & \textbf{0.36$\pm$0.19} & \textbf{81.60\%} & \textbf{3.89$\pm$1.31} \\
        \Xhline{3\arrayrulewidth}
    \end{tabular}
    \caption{S2 Results. For each measure, the best performing condition is bolded.}
    \label{table:S2_results}
\end{table*}

Similar to what we have observed in S1, there were no significant differences between conditions in terms of M1: prediction accuracy performance measure. On the contrary, attention quality-related measures all presented significant differences between conditions. 

\begin{table*}[b!]
    \centering
    \small
    \vspace{-3mm}
    \begin{tabular}{>{\raggedright\arraybackslash}p{32mm}|p{32mm}p{32mm}>{\raggedright\arraybackslash}p{32mm}p{32mm}p{32mm}>{\raggedright\arraybackslash}p{32mm}>{\raggedright\arraybackslash}p{32mm}}
        \Xhline{3\arrayrulewidth}
        \textbf{Condition Pairs} & \textbf{M3. IoU} & \textbf{M4. Attention Accuracy} & \textbf{M5. Perceived Quality}
        \\ 
        \Xhline{2\arrayrulewidth}
        \textbf{C1 vs. C2} & \textbf{1.57e-24}\textsuperscript{\textdaggerdbl\textdaggerdbl} & \textbf{7.31e-12}\textsuperscript{\textdaggerdbl\textdaggerdbl} & \textbf{2.47e-07}\textsuperscript{\textdaggerdbl\textdaggerdbl}
        \\
        \textbf{C1 vs. C3} & \textbf{1.48e-59}\textsuperscript{\textdaggerdbl\textdaggerdbl} & \textbf{1.20e-122}\textsuperscript{\textdaggerdbl\textdaggerdbl} & \textbf{7.91e-213}\textsuperscript{\textdaggerdbl\textdaggerdbl}
        \\
        \textbf{C2 vs. C3} & \textbf{1.43e-09}\textsuperscript{\textdaggerdbl\textdaggerdbl} & \textbf{4.94e-203}\textsuperscript{\textdaggerdbl\textdaggerdbl} & \textbf{1.89e-288}\textsuperscript{\textdaggerdbl\textdaggerdbl}
        \\
        \Xhline{3\arrayrulewidth}
    \end{tabular}
    \linebreak(\textdagger: $p < 0.05$, \textdaggerdbl: $p < 0.01$, \textdaggerdbl\textdaggerdbl: $p < 0.001$)
    \vspace{-3mm}
\end{table*}
\begin{figure*}[b!]
    \centering
    \includegraphics[width=\textwidth]{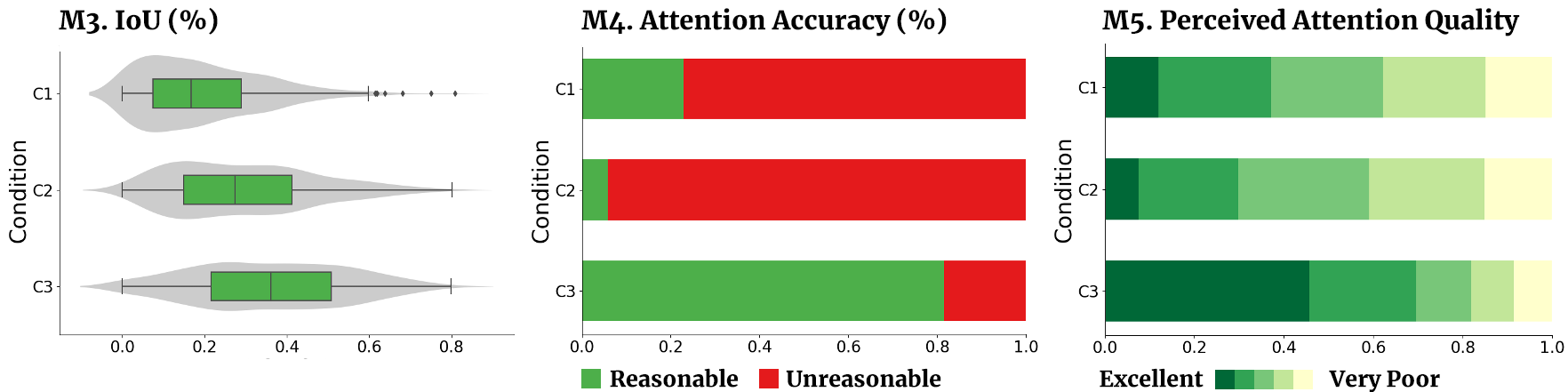}
    \caption{The pairwise comparison results of M3, M4, and M5 (Top) and distribution charts (Bottom)}
    \label{fig:S2_results}
\end{figure*}

Computation-based attention quality measure of M3 showed C3 \technique the best performing condition, followed by C2 ABN, and C1 Baseline. Since the distribution of IoU didn't follow the normal distribution, we applied Kruskal-Wallis H-test. The test found that there is a significant difference between the three conditions with a p-value of 1.67e-59. Dunn's test found the p-values of all pairs are below 0.05 (see M3 related metric in Fig.~\ref{fig:S2_results}).

Interestingly, human assessment-based measures of M4 and M5 disagreed with M3 in terms of the attention quality ranking. Same as M3, C3 turned out to be the best performing condition. However, both M4 and M5 showed that C1 outperforms C2. And the gaps between the three conditions were all significant. In terms of M4, The Kruskal-Wallis H-test shows a significant difference in the Attention Accuracy Performance of the three conditions with a p-value of 1.22e-221. The Dunn's test results for the post-hoc pairwise comparison are at the top in Fig.~\ref{fig:S2_results}. With the number of observations collected, the power analysis result shows a sufficient amount of power (all above 80\%) in our data to detect any significant difference between the conditions’ M4 performance. In addition, three annotators have substantial agreement with kappa coefficients that are at least 0.65. The M5, the perceived attention quality, followed the pattern we observed in M4. The Kruskal-Wallis H-test found there was a significance (p<0.05). The post-hoc comparison indicates that the condition's qualities are significantly different, pairwise, with p-values less than 0.05 (see Fig.~\ref{fig:S2_results}, M5). The pairwise effect sizes for M5 are: C1 vs. C2 (0.12), C1 vs. C2 (-0.72), and C2 vs. C3 (-0.87). So, any condition pair with C3 had a large negative effect size. By conducting the post-hoc power analysis, even with a relatively small effect size between C1 and C2, the number of observations was large enough to maintain the power (above 80\%) for further statistical tests.

\subsection{S2-2: Assessment through Model Performance Increase}
\subsubsection{Methodology}
In this subsection, we studied how the DNN models can benefit from IAA to gain a better generalization power under the scenarios where the training samples are limited. Specifically, we randomly sample a certain amount of training samples from our original validation set pool and provide the human annotation labels for each selected sample. Next, we use the samples as well as the attention labels as the training set to fine-tune a pretrained ResNet18, and evaluate the model performance with ROC-AUC score on the original test set. Similar to the previous study, we made comparative studies among the following three models: 1) Baseline, 2) ResNet18 + ABN, and 3) \technique. Notice that this setting does not require the reasonability assessment like the previous studies. Instead, here we assume that all samples have attention labels available, as we have enough human resources under the scenarios where the training samples are limited.

\subsubsection{Results}
We simulated four limited training sample scenarios, i.e. 1-shot, 5-shot, 10-shot, and 50-shot, where the number of shots means the total number of training samples per class. For example, in our gender classification dataset, 5-shot means we sample a total of 10 image samples, 5 for male class and 5 for female class.

\begin{figure}[!t]
\centering
\begin{subfigure}{.5\textwidth}
  \centering
  \includegraphics[width=\linewidth]{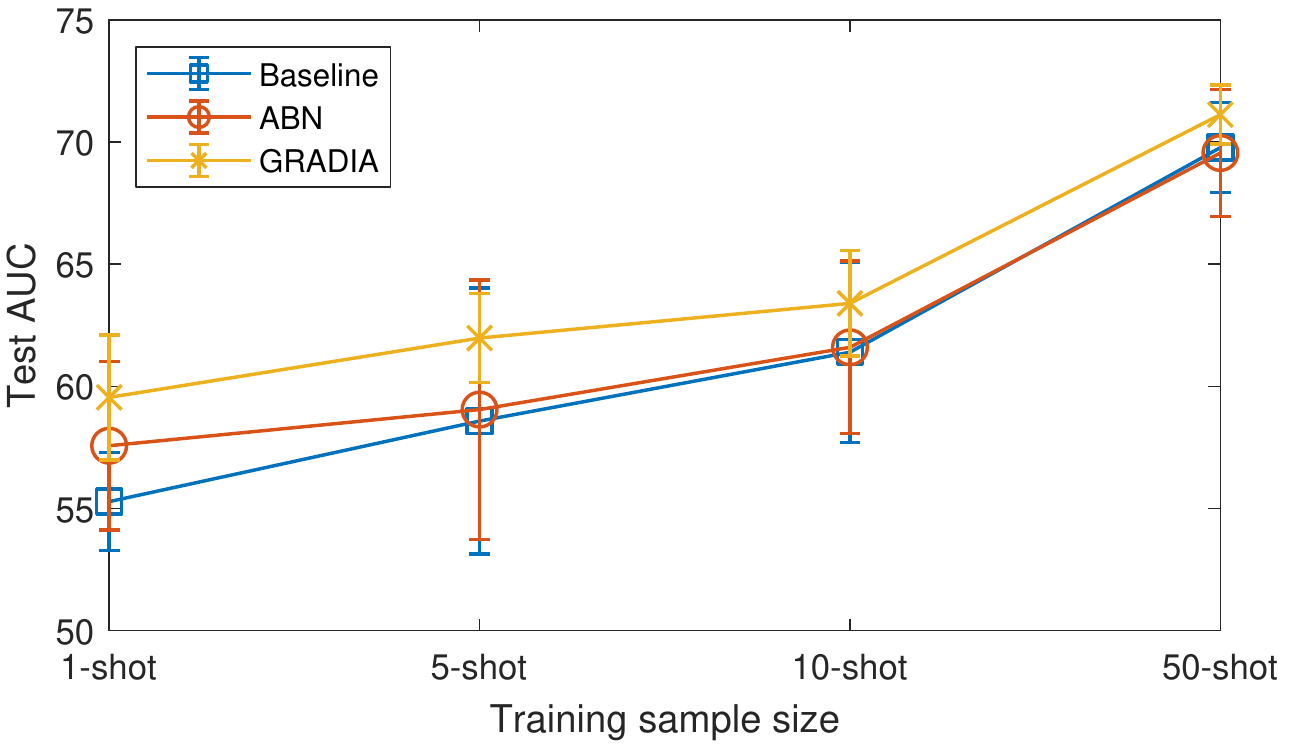}
  \caption{}
  \label{fig:trend}
\end{subfigure}%
\begin{subfigure}{.5\textwidth}
  \centering
  \includegraphics[width=\linewidth]{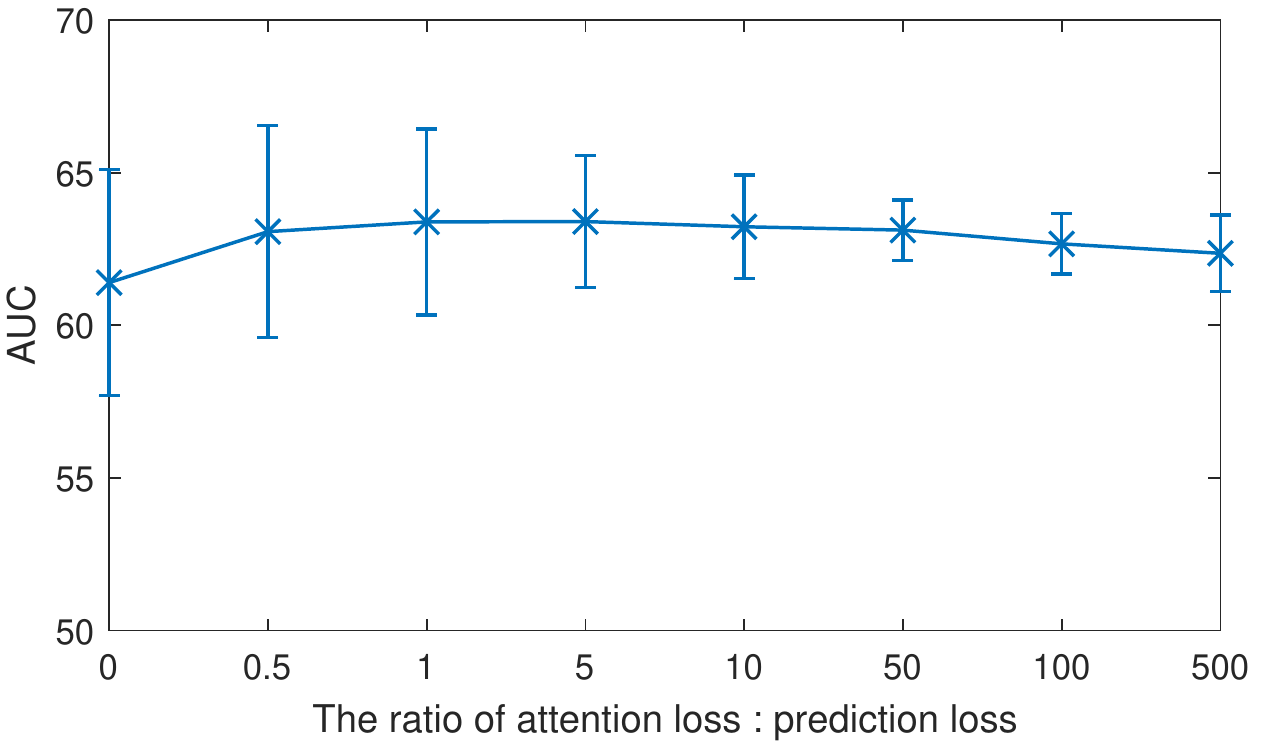}
  \caption{}
  \label{fig:sensitivity}
\end{subfigure}
\caption{Model performance under limited training samples. The data point represents the mean value over 10 random sample selection seeds, and the error bar here corresponds to the standard deviation. (a) The test AUC score comparison under different training sample size scenarios. (b) The sensitivity study of GRADIA under different ratio of perdition and attention loss on 10-shot scenario.}
\label{fig:S2_few}
\end{figure}

As shown in Fig.~\ref{fig:S2_few} (a), we present the test AUC score under the four training sample size scenarios. 
The data point here represents the mean value of the test AUC scores over 10 random training sample selection seeds, and the error bar here corresponds to the standard deviation. 
We can see that the proposed \technique outperforms all baseline models by a significant margin under all scenarios studied.
Specifically, \technique is able to improve the baseline model performance by 7.7\%, 5.8\%, 3.3\%, and 1.9\%, respectively under 1-shot, 5-shot, 10-shot, and 50-shot scenarios. 
Notice that ABN could also improve the model performance by leveraging the additional human attention labels, but generally much less effective than \technique. This is largely due to the additional layers and model parameters that are required in building the attention branch in ABN. This, on one hand, requires a large number of samples to learn how to generate attention, and on the other hand, makes the model more complex and prone to overfitting under a small training sample size.

We also studied how will different balance factors between the prediction and attention loss affect the final model performance. Fig.~\ref{fig:S2_few} (b) shows the sensitivity study of \technique under different ratios of perdition and attention loss on the 10-shot scenario. We can obtain two major findings: 1) the variance of the AUC score will be reduced as more weights are put to the attention loss; 2) the improvement of model performance is not very sensitive to the choice of the balance factor, as long as it is not set to 0 (which means we don't use the attention loss, thus this is equivalent to the baseline model). Those observations justified the general effectiveness of \technique under the scenarios where we don't have enough samples to train a model. Thus, our study can benefit the application domains where large amount of labeled data are difficult to obtain, such as in the weakly supervised learning scenarios~\cite{zhou2018brief}.

\subsection{Discussion}

\begin{figure*}[!t]
    \centering
    \includegraphics[width=\textwidth]{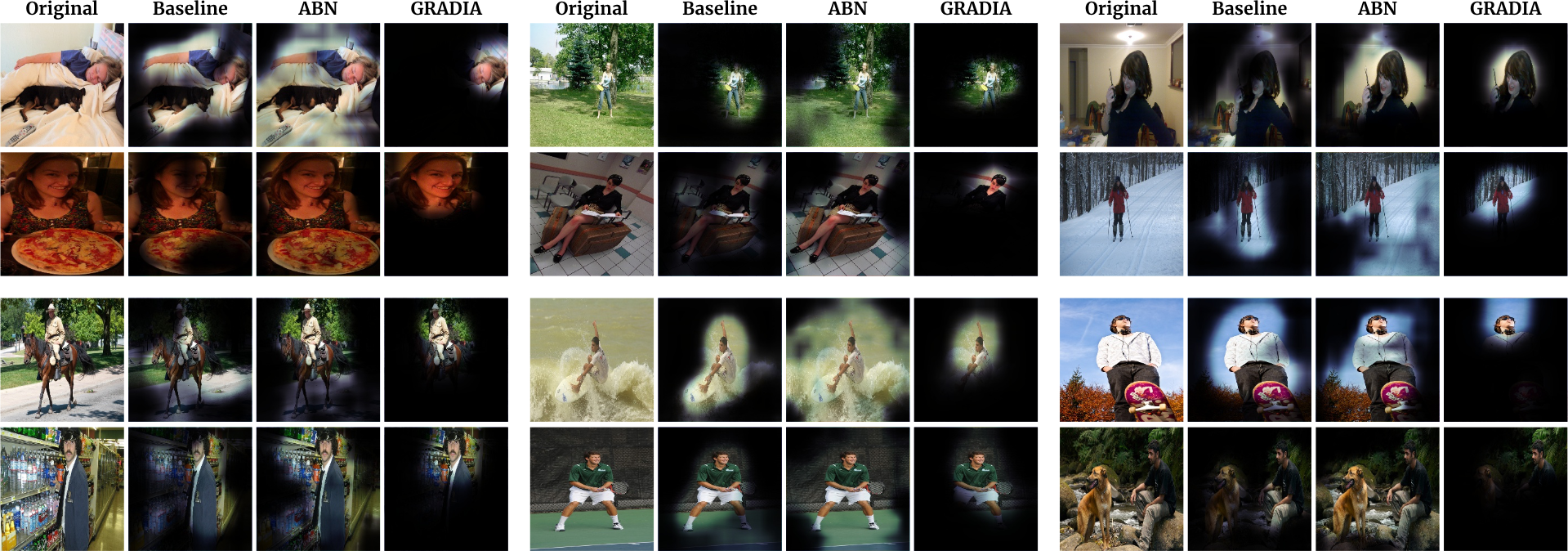}
        \caption{Twelve sets of attention maps that show how attention varies across the three conditions}
    \label{fig:S2_results2}   
    \vspace{-5mm}
\end{figure*}

In general, fine-tuning DNNs using \technique showed clear improvement on the quality of attention side without penalizing the prediction accuracy performance. Computational and human-assessment-based measures all indicated this pattern.  Across the condition, we did not observe the ``accuracy-fairness gap'' in this study again. Compared to C1, C3's Reasonability matrix showed instances categorized into Reasonable Accurate growing more than 3 times (147 to 515) while Unreasonable shrink more than one third. Like what we've observed, Reasonable Inaccurate increased from 25 to 97, showing the side effect of having ``concentrated gaze''. 

What made us rather surprising was the results of C2. Although M3 showed significantly improved IoU than C1, human-based assessments showed significantly lower ratings compared to C1. In order to understand such disagreement between computational and human-assessment-based measures, we posed every attention map results and checked how the gazes vary between the three conditions. As a result, we found two notable patterns. First, ABN tends to add more attention weight on the areas that necessary to identify gender, which would likely elicit ``yes'' on the first question in the reasonability questionnaire (asks whether the attention includes sufficient detail for performing a gender classification: Q1, hereinafter). Indeed, C2 had the highest number of yes in Q1 (n=739 out of 750) than C1 (n=725) and C3 (n=670). However, at the same time, we found the second pattern that the attention would likely to disperse in C2 to include more ``contextual objects'' while C3 strictly concentrates its gaze on intrinsic objects. We assume this aspect of ``dispersing gaze'' of ABN made human annotators answering the second reasonability question (Q2, hereinafter) also ``yes'' (meaning they perceive contextual object(s) not directly related to classifying gender). For C1, 706 instances were rated they include contextual objects in their attention maps while C1 had 576 C3 had only 112. This aspect influenced C2's reasonability matrix to include a huge amount of Unreasonable Accurate samples (n=580) which is worse than baseline (n=462). 

This misalignment between computational and human-assessment-based measures implies the necessity of diversifying the way we understand the DNN's performance, especially when the model's task encompasses context-dependent and subtle human subjectivity~\cite{chung2021understanding}. When Mitsuhara et al. evaluated the ABN in their study, they used the ``deletion metric'' which measures the decrease of prediction accuracy score change by gradually deleting the high attention area of an attention map~\cite{mitsuhara2019embedding}. Such an approach may result in a model with high accuracy, but this doesn't guarantee that model made a prediction with a ``right'' reason~\cite{hendricks2018women}. By including humans in-the-loop of evaluation, we argue that we can possibly design/build the model trustworthy, reasonable, while high-performing.

Besides, in our few sample learning studies, we observed that \technique can improve both model explanation quality as well as predictive power with the additional supervision on the model's attention. 
As nowadays DNNs are known to be data-hungry which requires a large number of training samples to work well, we hope our new findings here can provide some insights and motivate future studies on the directions towards learning more reliable and interpretable DNNs with much less amount of labeled data than in current stage.
\section{Implications for Design}

Through the two studies, we observed the quality of attention changes depending on (1) how we select the instances that need their attention to be adjusted and (2) which technique we use in leveraging the adjusted attention in the fine-tuning process. This section introduces aspects that one can consider when deciding to apply our framework or conducting research on \framework for steerable DNNs.

\textbf{Defining a Task}: In applying the IAM, we chose a problem of gender classification to demonstrate the steering effect of our approach. More fundamentally, however, what our mechanism aims at achieving is empowering humans to decide good and bad attention types and emphasize or dwindle those as wish. This can allow our framework to be applied in different application areas than the task we used in this work. For example, consider the case for building DNNs for classifying a human emotion. Merely focusing on a human face may not result in an accurate prediction when her expression is neutral but her background shows she is winning a medal in international competition~\cite{kosti2017emotic}. Contextual objects become indispensable assets in such cases, which is an opposite setting than ``de-bias''. In the domain of Visual Question Answering, ``common-sense'' displayed in images both introduces context and bias~\cite{zellers2019recognition} which requires a human to decide whether one will augment the influence of context or not. But regardless of any directions, a user of IAM must have the capability to decide whether (s)he would augment or kill the influence of context by indicating the desired focus in interactive editing. We note that although this direction is theoretically feasible, we have little evidence yet, and validation of our approach in different applications requires further studies.

\textbf{Elicitation of Human Annotation}: In order to use IAM, we leverage human annotators in two phases: when (1) structuring Reasonability Matrix and (2) eliciting attention adjustment. Once we decide on the task, next problem is to make the right question that can validate the reasonability of the instance. When developing questions, we suggest breaking down a question of ``Is the gaze given to the image reasonable?'' to more specific, possibly multiple questions (e.g., we broke down the reasonability question to two we specified in Reasonability Questionnaires). In that way, we can expect high-quality elicitation that could lead to better attention adjustment. One side note we wanted to mention is the high cost of eliciting annotation from human labelers. We expected asking reasonability questions can be a lightweight task, but it turned out to be not always the case. The elicitation process led us to spend more time than we expected at the beginning. Not a single participant answered within a day. Fast annotator finished it in three days and in the longest case, it extended to a week. We believe one of the imminent problem to be solved to increase the applicability of IAM is to build an interactive platform that a single human can specify the rule-based query (e.g., select the attention that includes things other than humans) to browse inferior attentions and batch-adjust multiple attentions at once such that even a few humans can handle thousands of images at once.

\textbf{Fine-tuning DNNs}: Based on our experimental study, \technique shows the most desirable results from both computers and humans. When designing fine-tuning mechanism, we suggest using our technique. One crucial shortcoming in our approach is its interactivity. Although we named our framework as ``Interactive'' Attention Alignment, the way we used our framework was not interactive, as even if we finish adjusted the attention, seeing the effect of adjusted attention made using a test set is not real-time. Even though fine-tuning is a computationally expensive procedure that requires some time, improvement of round-trip speed should be necessary. Align with this aspect, the current approach is not presented in a visual analytic platform which could introduce some gap to users who are not familiar with terminal-based interfaces and hidden settings. In the future, we hope there can be commitment towards developing visual analytic tools that a user can easily specify the split ratio, intelligently browse reasonable/unreasonable instances, performing batch adjustment, and seeing the aftermath of adjustment in a faster round trip speed (ideally in an interactive manner).

\textbf{Beyond Convolutional Neural Network (CNN) and Image Data}: Lastly, our approach has been built based on CNNs which operate on image data. 
Since the IAM framework can be general and model-agnostic, we believe the proposed framework can also be easily extended to other data types and the corresponding DNN models. 
For example, one of our possible future work is to apply IAM to steer the explanation of Graph Convolutional Networks (GCN)~\cite{kipf2016semi} on the graph-structured data such as sense graphs and molecule graphs~\cite{pope2019explainability}.
In all, we believe the direction of study on IAM can be beneficial to help DNNs to better align their attention/explanation with human and consequentially enhance our understanding of the machine learning models as a whole.

\section{Conclusion}
As the use of DNNs is becoming pervasive in many critical domains, people's focus on building an accurately performing model has rapidly shifted to understanding how DNNs work under the hood and more importantly, how to adjust the way DNNs work based on human knowledge, experience, and expectation. The overarching motivation behind this work is to devise a novel interaction modality that a human can leverage in steering DNNs in a direct, intuitive, and intelligible manner. To do so, this work aimed at laying the groundwork towards establishing a platform that can use \framework to more directly infusing their perspectives in fine-tuning DNNs. As a closing remark, we hope this work can motivate future research in \framework on DNN and more generally devising novel interaction modalities that can and realize DNNs that align with a human mental model.
\section{Acknowledgments}
This work was supported by the NSF Grant No. 1755850, No. 1841520, No. 2007716, No. 2007976, No. 1942594, No. 1907805, a Jeffress Memorial Trust Award, Amazon Research Award, NVIDIA GPU Grant, and Design Knowledge Company (subcontract number: 10827.002.120.04).

\bibliographystyle{ACM-Reference-Format}
\bibliography{main}

\end{document}